\documentclass{article}
\usepackage{ijcai18}
\usepackage{epsfig}
\usepackage{graphicx}
\usepackage{amsmath}
\usepackage{graphicx}
\usepackage{amssymb}
\usepackage{float}
\usepackage{multirow}
\usepackage{graphicx}
\usepackage{xr}

\usepackage{times}
\usepackage{xcolor}
\usepackage{soul}
\usepackage[utf8]{inputenc}
\usepackage[small]{caption}

\title{Long-Term Human Motion Prediction by Modeling Motion Context and Enhancing Motion Dynamic}

\author{
Yongyi Tang$^{1}$\thanks{Work done while Yongyi Tang was a Research Intern with Tencent AI Lab.}\qquad Lin Ma$^2$\renewcommand{\thefootnote}{\fnsymbol{footnote}}\footnotemark[2] \qquad Wei Liu$^2$ \qquad Wei-Shi Zheng{$^3$}\renewcommand{\thefootnote}{\fnsymbol{footnote}}\footnotemark[2]
\\ 
$^1$School of Electronics and Information Technology, Sun Yat-sen University\\
$^2$Tencent AI Lab\\
$^3$School of Data and Computer Science, Sun Yat-sen University\\
\small{\texttt{\{yongyi.tang92, forest.linma, wliu.cu\}@gmail.com \qquad wszheng@ieee.org}}
}

\begin{document}

\maketitle
\renewcommand{\thefootnote}{\fnsymbol{footnote}}
\footnotetext[2]{Corresponding authors.}

\begin{abstract}
Human motion prediction aims at generating future frames of human motion based on an observed sequence of skeletons. 
Recent methods employ the latest hidden states of a recurrent neural network (RNN) to encode the historical skeletons, which can only address short-term prediction. 
In this work, we propose a motion context modeling by summarizing the historical human motion with respect to the current prediction. 
A modified highway unit (MHU) is proposed for efficiently eliminating motionless joints and estimating next pose given the motion context. 
Furthermore, we enhance the motion dynamic by minimizing the gram matrix loss for long-term motion prediction. 
Experimental results show that the proposed model can promisingly forecast the human future movements, which yields superior performances over related state-of-the-art approaches.
Moreover, specifying the motion context with the activity labels enables our model to perform human motion transfer.

\end{abstract}


\externaldocument{Experiments}
\section{Introduction}

\begin{figure}[ht]
\begin{center}
   \includegraphics[width=1\linewidth]{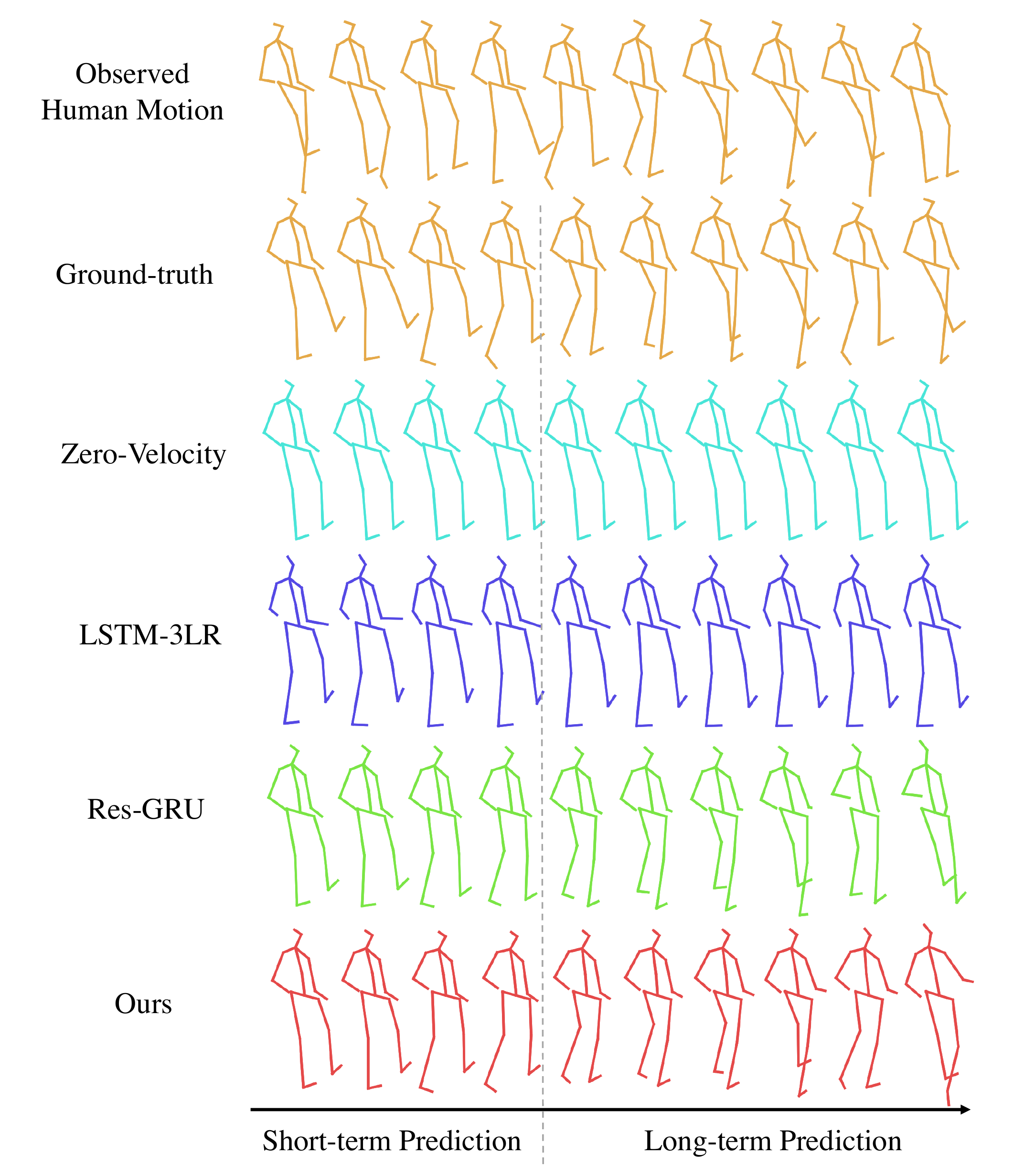}
\end{center}
\vspace{-11pt}
   \centering\caption{Human motion prediction on ``walking''. Top: the observed human motion sequence. Given the observed skeletons, the goal of this paper is to generate future skeletons similar to the ground-truth (the second row). Our method is able to well predict both short-term and long-term skeletons maintaining good temporal dynamic, while other existing methods fail to generate satisfying long-term skeletons. Better view in color.}
\label{fig:mot}
\vspace{-15pt}
\end{figure}

Human motion prediction, serving as one of the most essential parts of robotic intelligence, enables rapid and high-fidelity reactions towards complex environment changes. 
For example, a robot can effortlessly avoid route collision by forecasting the movement about surrounding subjects.
Nowadays,
with the development of the MOCAP devices, such as Kinect, and the pose estimation algorithms \cite{yasin2016dual,tekin2017learning}, the sequence of human skeletons can be easily and accurately computed. It thus enables us to predict future human motion by analyzing the observed skeleton sequences, which can further help human action analysis/recognition, body pose estimation, and even human-robot interactions.

The historical human skeleton sequence needs to be effectively modeled for human motion prediction \cite{fragkiadaki2015recurrent,jain2016structural} and action recognition \cite{wang2014learning,liu2016spatio}. Currently, one common strategy is to use a recurrent neural network (RNN) as the encoder along the temporal domain \cite{fragkiadaki2015recurrent,Ghosh2017learning,martinez2017human}, driven from sequence to sequence learning \cite{sutskever2014sequence}, with the last hidden state encoding the motion context. For the existing recurrent units such as long short-term memory (LSTM) \cite{hochreiter1997long} and gated recurrent unit (GRU) \cite{cho2014learning}, the hidden states encode the skeleton sequence and update at every time step. Although LSTM and GRU are proposed to handle the long short-term dependencies, the historical information, especially the long term one, cannot be well encoded with the updated hidden states overwhelmed by the input at current step \cite{bahdanau2014neural}. 
Such information loss makes the long-term human motion prediction tend to converge to the mean pose or fail to produce motion dynamic, as the results of LSTM-3LR \cite{fragkiadaki2015recurrent} and Res-GRU \cite{martinez2017human} shown in Fig.~\ref{fig:mot}. 
Moreover, the human joints of skeleton are treated equally for the motion prediction in the prior works. 
Instead, the human motion can be viewed as the movement of the joints of the skeleton, where not every joint participates in the human pose evolutions. For the human activities, such as ``walking'' and ``eating'', the subject may stand still with the backbones motionless.


In order to make reliable future predictions, we model motion context by summarizing the historical human motion skeleton sequence with respect to the current skeleton. Such motion context can help to capture the human motion patterns, $i.e.$ the repeated patterns in ``walking'' and ``eating'', and ease the motion uncertainties, thus benefiting the long-term predictions.
By utilizing both the pose information of the last frame and the summarized motion context, we propose a modified highway unit (MHU) to predict the future human skeleton. The MHU introduces a gate that can efficiently filter the motionless joints at each generation and pay more attentions on those with motion.
Besides, in order to produce consistent human motions and enhance the motion dynamic, we introduce a gram matrix loss for minimization so as to explicitly penalize the mean pose convergence and ease error accumulation.
These components enable our method to predict reliable long-term human motion as highlighted in the last row of Fig.~\ref{fig:mot}.

In addition, prior works are only able to predict one single activity for a given pose sequence, $i.e.$ predicting future ``walking'' skeletons given the observed ``walking'' skeleton sequence. However, in realistic scenario, more than one type of activity may evolve given the observed human motion sequence.
With our motion context modeling, we further exploit the ability of the proposed model on human motion transfer, which generates specific types of motion sequence given different action labels. 
As such, the human motion sequence can be manipulated by the given action labels, resulting in a smooth motion sequence with multiple activities, which will be detailed in Sec.~\ref{experiments}.

Our contributions are summarized as follows:
1) We propose to model the motion context by summarizing the historical skeleton sequence with respect to the current one. MHU thereafter distinguishes the motion joints from the motionless ones to make effectively long-term human motion prediction.
2) A gram matrix loss is proposed for enhancing motion dynamic, which enables our model to produce highly correlated human motion in the temporal domain.
3) Our model can perform human motion transfer based on the motion context and the specified activity categories.


\section{Related Works}
\noindent \textbf{Human Motion Analysis.} 
Human motion analysis is one of the key problems in computer vision and robotics, and hence has received much attention \cite{aggarwal1997human}. 
Human motion can be obtained by motion capture systems \cite{h36m_pami} and Kinect device and extracted from videos \cite{brand2000style} and even static images \cite{li20143d,yasin2016dual}.
With the available body poses, several structural models such as hierarchical recurrent neural networks \cite{du2015hierarchical} and trust gates \cite{liu2016spatio} were proposed to address skeleton based action recognition. 
By representing skeletons with the rotation matrices, which forms a special orthogonal group SO(3), the researches in \cite{vemulapalli2014human} and \cite{huang2017deep} developed group-based skeleton analysis for action recognition by using support vector machine and convolution neural network, respectively.

\noindent \textbf{Human Motion Prediction.} 
Human motion prediction aims to understand behaviors of a subject on the observed sequences and to generate future body poses.
Deep learning based approaches have outperformed conventional methods on skeleton-based problem such as 3D pose estimation \cite{yasin2016dual} and action recognition \cite{hu2015jointly,liu2016spatio}. In this paper, we focus on human motion prediction based on deep neural networks.
Prior works try to encode the observed information to latent variables and perform prediction as decoding by Restricted Boltzmann Machines (RBMs) \cite{taylor2007modeling}. 
\cite{fragkiadaki2015recurrent} introduce Encoder-Recurrent-Decoder networks that learn the temporal dynamic of human motion by a long short-term memory (LSTM) model. They designed a non-linear transformation to encode pose feature and decode the output of the LSTM. The history information passes throughout the recurrent units to constrain human motion prediction.
\cite{martinez2017human} further extended this scheme by modeling the velocity of joints instead of directly estimating the body pose, and employed single linear layer for pose features encoding and hidden states decoding. They find that the poses with zero-velocity achieve relatively less error on mean angle distance, which demonstrates the efficiency of the velocity modeling.
To reduce the accumulated correlation error, a dropout auto-encoder (DAE) was proposed by \cite{Ghosh2017learning}. 
Apart from these approaches, structural RNN proposed by \cite{jain2016structural} tries to capture the spatio-temporal relationship of joints.

\begin{figure*}[t]
\begin{center}
   \includegraphics[width=0.9\linewidth]{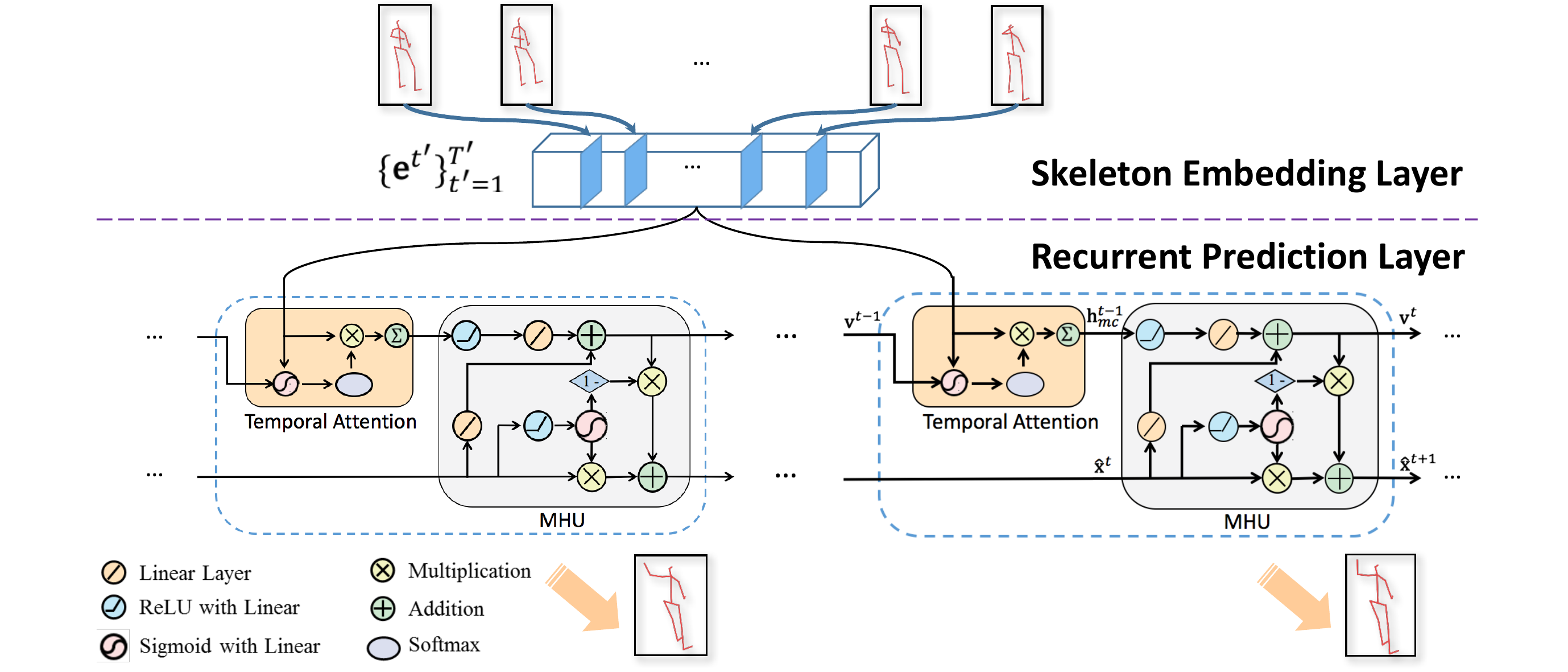}
\end{center}
\vspace{-11pt}
   \centering\caption{The architecture of our proposed model for human motion prediction. Each historical skeleton is first embedded into one semantic space. At each time step, the motion context modeling summarizes the skeleton embeddings with respect to the last predicted skeleton. Afterwards, MHU works on the motion context and the last estimated skeleton to yield the human motion at each time step. }
\label{fig:framework}
\vspace{-11pt}
\end{figure*}

\section{Proposed Model}
\subsection{Problem Formulation}
Given an observed sequence of body poses $\{\mathbf{x}^{t'}\}^{T'}_{t'=1}$ in 3D space, the goal of human motion prediction is to generate the consecutive human motion $\{\hat{\mathbf{x}}^{t}\}^{T}_{t=1}$ close to the ground-truths $\{\mathbf{x}^{t}\}^{T}_{t=1}$. 
Following the prior works \cite{fragkiadaki2015recurrent,martinez2017human} for human motion prediction, the axis-angle representation of skeletons $\{\mathbf{x}^{t'}\}^{T'}_{t'=1}$ parameterizes a rotation of each joint in a three-dimensional Euclidean space by a rotation vector whose norm is the rotation angle.

Conventional RNN-based human motion prediction methods \cite{fragkiadaki2015recurrent,jain2016structural,martinez2017human} rely on the last hidden state $\mathbf{h}^{t-1}$ and predicted skeleton $\hat{\mathbf{x}}^{t}$:
\begin{equation}
\hat{\mathbf{x}}^{t+1} = \text{RNN}(\hat{\mathbf{x}}^{t}, \mathbf{h}^{t-1}).
\end{equation}
The historical human skeletons are encoded by $\mathbf{h}^0$ for predicting the first skeleton $\hat{\mathbf{x}}^{1}$. 
However, the failure cases in long-term human motion prediction (as shown in Fig.~\ref{fig:mot}) indicate that using the final hidden state as the motion context is not satisfactory to well capture the historical motion information.

To address the above problems, we aim at designing a model $f$ equipped with motion context modeling to fully explore the properties of human motion sequences, which is further formulated as:
\begin{equation}
\hat{\mathbf{x}}^{t+1} = f(\hat{\mathbf{x}}^{t}, \{\mathbf{x}^{t'}\}^{T'}_{t'=1}).
\end{equation}
Our proposed model $f$ directly accesses to the historical human skeletons $\{\mathbf{x}^{t'}\}^{T'}_{t'=1}$ at each step for prediction, which enables us to yield a more representative motion context. As such, the model can simply repeat the observed pattern to get a reasonable prediction for periodic activities such as ``walking'' and ``eating''. For aperiodic activities, the encoded motion context can still provide meaningful information of historical activities (such as directions or the habit of movement), and thus further reduce the search space for making predictions.

\subsection{Our Approach}

As shown in Fig.\ref{fig:framework}, the proposed model mainly consists of two components: a skeleton embedding layer and a recurrent prediction layer. The embedding layer can be regarded as an encoder, and the recurrent prediction layer is thus denoted as the decoder, which consists of two main components, namely the motion context modeling and the modified highway unit (MHU). These two components are coupled together, and this enables the proposed framework to predict reliable long-term human motions.

A multi-layer non-linear network is constructed to realize the skeleton embedding layer, which projects the observed skeletons $\{\mathbf{x}^{t'}\}^{T'}_{t'=1}$ into the semantic space yielding $\{\mathbf{e}^{t'}\}^{T'}_{t'=1}$.
Specifically, we concatenate the output of a fully connected layer $\mathbf{h}_{e1} = \mathbf{W}_{e1}\mathbf{x}^{t'}+\mathbf{b}_{e1}$ and its activated output $\mathbf{h}_{e2} = \mathrm{ReLU}(\mathbf{h}_{e1})$, and finally preform embedding by $\mathbf{e}^{t'} = \mathbf{W}_{e2}[\mathbf{h}_{e1};\mathbf{h}_{e2}]+\mathbf{b}_{e2}$.

During the prediction, the motion context is firstly summarized from the skeleton embeddings with respect to the last predicted skeleton. Afterwards, the MHU exploits the relationships between the motion context and the predicted skeleton to generate the human motion at each time step.

\subsubsection{Motion Context Modeling}
Motion context modeling aims at encoding the historical human motion, which can further boost the future skeleton prediction. 
Existing methods model motion context simply by LSTM or GRU, and the last hidden state is taken as motion context for human motion prediction  \cite{fragkiadaki2015recurrent,martinez2017human,zimo2017auto} and action recognition \cite{liu2016spatio,du2015hierarchical}. 
However, the last hidden state in RNN is usually dominated by the input at the latest time step. 
Therefore, the previous information, especially for the long-term one, is not effectively encoded into the hidden state.
While for future motion prediction, the historical skeletons are believed to be helpful.

In this paper, we propose to use temporal attention mechanism \cite{bahdanau2014neural} to summarize all the historical skeletons with the respect to predicted one at each time step:
\begin{equation}
\beta^{t'} = \mathbf{W}_{\beta}\text{tanh}(\mathbf{U}_{\beta v}\mathbf{v}^{t-1} + \mathbf{U}_{\beta e}\mathbf{e}^{t'} + b_{\beta}) ,
\end{equation}
\begin{equation}
\alpha^{t'} = \frac{\text{exp}({\beta^{t'}})}{\sum^{T'}_{t'=1}\text{exp}({\beta^{t'}})},
\end{equation}
where $\mathbf{v}^{t-1}$ denotes the predicted skeleton at time $t-1$. $\alpha^{t'}$ denotes the attentive weight with respect to each historical skeleton. The temporal attention mechanism directly works on the skeleton embeddings, which can more effectively capture the relations between the predicted skeleton and historical motion. With the computed attentive weights, the motion context is thus computed by:
\begin{equation}
\mathbf{h}_{mc}^{t-1} = \sum^{T'}_{t'=1}\alpha^{t'}\mathbf{e}^{t'}.
\end{equation}

The obtained motion context $\mathbf{h}_{mc}^{t-1}$ can selectively summarize the historical skeleton information. The obtained motion context presents no bias on short-term or long-term information. Thus, $\mathbf{h}_{mc}^{t-1}$ can help produce more reliable long-term predictions compared with the state-of-the-art methods which directly use the last hidden state of traditional RNNs.

\subsubsection{Modified Highway Unit}
The human motion can be viewed as the movement of skeleton joints, where not every joint participates in pose evolutions. For example, one subject mainly stands still with the backbones presenting motionless in activities such as ``phoning'' and ``eating''. Therefore, the human motion is only triggered by the activity-specific skeleton joints.

Based on these observations, we introduce an MHU in our recurrent prediction layer as shown in Fig.\ref{fig:framework} in order to efficiently model the skeleton joints that contain meaningful motion information.
We introduce ReLU non-linearity in the Recurrent Highway Network proposed by \cite{zilly2016recurrent} for both skeleton estimation and gate estimation.
We additionally drop the $tanh$ activation in the vanilla RHN.

Given the current input skeleton representation $\mathbf{x}^{t}$ and the motion context $\mathbf{h}_{mc}^{t-1}$ from the last time step, our proposed MHU is formulated as:
\begin{equation}
\mathbf{v}^t = \mathbf{W}_{v}\phi(\mathbf{U}_{vh}\mathbf{h}_{mc}^{t-1}+\mathbf{b}_v) + \mathbf{U}_{vx}\mathbf{x}^t + \mathbf{b}_{vh} ,
\end{equation}
\begin{equation}
\mathbf{z}^t = \sigma(\mathbf{W}_{z}\phi(\mathbf{U}_{zx}\mathbf{x}^t+\mathbf{b}_z)+\mathbf{b}_{zx}) ,
\end{equation}
\begin{equation}\label{eq:eq8}
\hat{\mathbf{x}}^{t+1} = (\mathbf{1}-\mathbf{z}^t)\odot \mathbf{v}^t + \mathbf{z}^t\odot \mathbf{x}^t ,
\end{equation}
where $\odot$ is element-wise multiplication, $\mathbf{W}$, $\mathbf{U}$ and $b$ are the learned parameters, and $\phi$ and $\sigma$ denote the rectified linear unit and sigmoid function, respectively.

Note that $\mathbf{z}^t$ ranges in [0,1] for gating the current skeleton $\mathbf{x}^t$ and the estimated next joints $\mathbf{v}^t$. 
The modeling of gate state $\mathbf{z}^t$ involves a non-linear transformation, which implicitly captures the spatial relations of $\mathbf{x}^t$. As such, MHU is expected to focus on the joints with large motions, and conducts partially updating in Eq.\ref{eq:eq8} by the element-wise multiplication. These non-linear operations within the MHU can help explore spatial relations of skeleton joints.


\subsection{Enhancing Motion Dynamics with Gram Matrix Objective}

In addition to model motion context for long-term prediction, the transitions between skeletons should be addressed to generate dynamic human motion and prevent mean pose convergence.
To minimize the error of human motion prediction, existing methods \cite{fragkiadaki2015recurrent,jain2016structural,martinez2017human} usually adopt mean square error (MSE) as the objective function.
However, the MSE constrains models to generate the human motion that stays around the center of the ground-truth distribution, which are the mean poses. 
Moreover, the MSE objective only treats each motion independently which may cause motion inconsistency.
Instead, we propose to minimize the gram matrix between consecutive motions, which is defined as follow:
\begin{equation}
\mathcal{L}_{gram} = \frac{1}{T}\sum^{T-1}_{t=1}\big|\big|G(\hat{\mathbf{x}}^t,\hat{\mathbf{x}}^{t-1}) - G(\mathbf{x}^t, \mathbf{x}^{t-1})\big|\big|^2_2,
\end{equation}
where the gram matrix $G(\mathbf{x}^t, \mathbf{x}^{t-1})$ is defined as:
\begin{equation}
G(\mathbf{x}^t, \mathbf{x}^{t-1}) = [\mathbf{x}^t;\mathbf{x}^{t-1}][\mathbf{x}^t;\mathbf{x}^{t-1}]^\top,
\end{equation}
and $[\cdot;\cdot]$ denotes the concatenation of vectors.

On one hand, the correlation between skeleton joints is represented in the gram matrix such that the spatial relation among different skeleton joints can be further explored. 
On the other hand, the temporal dynamic is captured by the correlation between $\mathbf{x}^t$ and $\mathbf{x}^{t+1}$, which enables our model to enhance human motion along temporal axis. 
For the action such as ``walking'', the arms and legs move alternately. Such spatial-temporal correlation represented in the gram matrix can enable producing human-like walking motion.
Thus both short-term and long-term human motion predictions can be improved.

\section{Experiments}\label{experiments}
\subsection{Experimental Settings}
\noindent{\bf{H3.6m Mocap Dataset for Human Motion Prediction}}

\noindent We conducted our experiments of human motion prediction on the H3.6m mocap Dataset \cite{h36m_pami}, which is the largest human motion dataset for 3D body pose analysis.
It consists of 15 activities including periodic activities like ``walking'' and non-periodic activities such as ``discussion'' and ``taking photo'', performed by seven different professional actors. Recorded by a Vicon motion capture system, the H3.6m dataset provides high quality 3D body joint locations in the global coordinate sampled at 50 frames per second (fps).

\vspace{5pt}
\noindent{\bf{Data Representation and Preprocessing}}

\noindent For all our experiments, we followed the same data setting in \cite{fragkiadaki2015recurrent,jain2016structural,martinez2017human}.
The motion sequence was down-sampled by 2 to 25 fps. And 5 subjects were selected for testing with the others for training.
The joint features were represented in exponential map \cite{grassia1998practical} which is also known as the angle-axis representation. The three dimension feature of each joint represents the rotation vector with respect to the parent joint predefined in H3.6m dataset.
All the features were normalized into the range of [-1,1]. We did not use the label as additional information except for the experiments of human motion transfer. 

\vspace{5pt}
\noindent{\bf{Training}}

\noindent Single layer of MHU with 1024 units was adopted in all our experiments. Empirically, stacking more layers of MHU did not help improve the performance.
To better capture human motion, all the activities were trained together for prediction as a default setting.
We used $T'=30$ observed frames for embedding to estimate future $T=10$ frames.
We used stochastic gradient descent with the momentum setting to 0.9.
The learning rate was set to 0.05 decayed with factor of 0.95 for every 10,000 steps. And the gradient was clipped to a maximum L2-norm of 5.
Batch size of 80 was used throughout our experiments. Normally, the training converged in around 20,000 steps.

\begin{table*}[t]
\caption{Performance comparison between different methods in terms of both short-term and long-term human motion prediction of all 15 activities from H3.6m dataset via mean angle error. The best performance is highlighted in boldface.}
\vspace{-6pt}
\resizebox{1\textwidth}{!}{
\begin{tabular}{|c|c|c|c|c|c|c|c|c|c|c|c|c|c|}
\hline
\multirow{2}{*}{Methods} 
&\multicolumn{5}{c|}{Short-term} & \multicolumn{8}{c|}{Long-term} \\
\cline{2-14}
   & 80ms & 160ms & 240ms& 320ms & 400ms & 480ms & 560ms & 640ms & 720ms & 800ms & 880ms & 960ms & 1000ms  \\
\hline
ERD \cite{fragkiadaki2015recurrent} & 0.93 & 1.07 & 1.19 & 1.31 & 1.41 & 1.52 & 1.58 & 1.64 & 1.70 & 1.78 & 1.86 & 1.93 & 1.95  \\
\hline
LSTM-3LR \cite{fragkiadaki2015recurrent} & 0.87 & 0.93 & 1.06 & 1.19 & 1.30 & 1.41 & 1.49 & 1.55 & 1.62 & 1.70 & 1.79 & 1.86 & 1.89 \\
\hline
Res-GRU \cite{martinez2017human} & 0.40 & 0.72 & 0.92 & 1.09 & 1.23 & 1.36 & 1.45 & 1.52 & 1.59 & 1.68 & 1.77 & 1.85 & 1.89  \\
\hline
Zero-velocity & 0.40 & 0.71 & 0.90 & 1.07 & 1.20 & 1.32 & 1.42 & 1.50 & 1.57 & 1.66 & 1.75 & 1.82 & 1.85  \\ 
\hline
\hline
MHU-MSE & \textbf{0.39} & 0.69 & 0.88 & 1.04 & 1.17 & 1.30 & 1.40 & 1.49 & 1.57 & 1.67 & 1.77 & 1.86 & 1.89  \\
\hline
MHU-Gram & \textbf{0.39} & \textbf{0.68} & 0.86 & \textbf{1.01} & 1.14 & 1.26 & 1.35 & 1.43 & 1.50 & 1.60 & 1.70 & 1.79 & 1.82 \\
\hline
Ours &\textbf{ 0.39} & \textbf{0.68} & \textbf{0.85} & \textbf{1.01} & \textbf{1.13} & \textbf{1.25} & \textbf{1.34} & \textbf{1.42} & \textbf{1.49} & \textbf{1.59} & \textbf{1.69} & \textbf{1.77} & \textbf{1.80} \\
\hline
 \end{tabular}
 }
     \label{Tab:all_table}
\end{table*}

\begin{table*}[t]
\caption{Performance comparison between different methods in terms of both short-term and long-term human motion prediction via mean angle error for each individual activity from H3.6m dataset, including  ``\texttt{walking}'', ``\texttt{Greeting}'', ``\texttt{Walking Dog}'', ``\texttt{Discussion}'', ``\texttt{Posting}'' and ``\texttt{Taking Photo}''. The best performance is highlighted in boldface.}
\vspace{-6pt}
\resizebox{1\textwidth}{!}{
\begin{tabular}{|c|c|c|c|c|c|c|c|c|c|c|c|c|c|c|c|c|}
\hline
\multirow{3}{*}{Methods} &
\multicolumn{8}{c|}{Walking} &
\multicolumn{8}{c|}{Greeting}\\
\cline{2-17}
&\multicolumn{4}{c|}{Short-term} & \multicolumn{4}{c|}{Long-term} & \multicolumn{4}{c|}{Short-term} & \multicolumn{4}{c|}{Long-term}\\
\cline{2-17}
   & 80ms & 160ms & 320ms & 400ms & 560ms & 640ms & 720ms & 1000ms & 80ms & 160ms & 320ms & 400ms & 560ms & 640ms & 720ms & 1000ms  \\
\hline
ERD \cite{fragkiadaki2015recurrent} & 0.77 & 0.90 & 1.12 & 1.25 & 1.44 & 1.45 & 1.46 & 1.44 & 0.85 & 1.09 & 1.45 & 1.64 & 1.93 & 1.89 & 1.92 & 1.98   \\
\hline
LSTM-3LR \cite{fragkiadaki2015recurrent} & 0.73 & 0.81 & 1.05 & 1.18 & 1.34 & 1.36 & 1.37 & 1.36 & 0.80 & 0.99 & 1.37 & 1.54 & 1.81 & 1.76 & 1.79 & 1.85 \\
\hline
Res-GRU \cite{martinez2017human} & \textbf{0.27} & \textbf{0.47} & \textbf{0.68} & \textbf{0.76} & \textbf{0.90} & \textbf{0.94} & 0.99 & \textbf{1.06} & \textbf{0.52} & \textbf{0.86} & 1.30 & 1.47 & 1.78 & 1.75 & 1.82 & 1.96 \\
\hline
Zero-velocity & 0.39 & 0.68 & 0.99 & 1.15 & 1.35 & 1.37 & 1.37 & 1.32 & 0.54 & 0.89 & 1.30 & 1.49 & 1.79 & 1.74 & 1.77 & \textbf{1.80} \\ 
\hline
Ours & 0.32 & 0.53 & 0.69 & 0.77 & \textbf{0.90} & \textbf{0.94} & \textbf{0.97} & \textbf{1.06} & 0.54 & 0.87 & \textbf{1.27} & \textbf{1.45} & \textbf{1.75} & \textbf{1.71} & \textbf{1.74} & 1.87 \\
\hline

\hline
\multirow{3}{*}{Methods} &
\multicolumn{8}{c|}{Walking Dog} &
\multicolumn{8}{c|}{Discussion}\\
\cline{2-17}
&\multicolumn{4}{c|}{Short-term} & \multicolumn{4}{c|}{Long-term} & \multicolumn{4}{c|}{Short-term} & \multicolumn{4}{c|}{Long-term}\\
\cline{2-17}
   & 80ms & 160ms & 320ms & 400ms & 560ms & 640ms & 720ms & 1000ms & 80ms & 160ms & 320ms & 400ms & 560ms & 640ms & 720ms & 1000ms  \\
\hline
ERD \cite{fragkiadaki2015recurrent} & 0.99 & 1.25 & 1.48 & 1.58 & 1.83 & 1.88 & 1.96 & 2.03 & 0.76 & 0.96 & 1.17 & 1.24 & 1.57 & 1.70 & 1.84 & 2.04 \\
\hline
LSTM-3LR \cite{fragkiadaki2015recurrent} & 0.91 & 1.07 & 1.39 & 1.53 & 1.81 & 1.85 & 1.90 & 2.00 & 0.71 & 0.84 & 1.02 & 1.11 & 1.49 & 1.62 & 1.76 & 1.99 \\
\hline
Res-GRU \cite{martinez2017human} & \textbf{0.56} & 0.95 & 1.33 & 1.48 & 1.78 & 1.81 & 1.88 & 1.96 & \textbf{0.31} & 0.69 & 1.03 & 1.12 & 1.52 & 1.61 & 1.70 & \textbf{1.87} \\
\hline
Zero-velocity & 0.60 & 0.98 & 1.36 & 1.50 & 1.74 & 1.80 & 1.87 & 1.96 & \textbf{0.31} & 0.67 & 0.97 & 1.04 & 1.41 & 1.56 & 1.71 & 1.96 \\
\hline
Ours & \textbf{0.56} & \textbf{0.88} & \textbf{1.21} & \textbf{1.37} & \textbf{1.67} & \textbf{1.72} & \textbf{1.81} & \textbf{1.90} & \textbf{0.31} & \textbf{0.66} & \textbf{0.93} & \textbf{1.00} & \textbf{1.37} & \textbf{1.51} & \textbf{1.66} & 1.88\\
\hline

\hline
\multirow{3}{*}{Methods} &
\multicolumn{8}{c|}{Posting} &
\multicolumn{8}{c|}{Taking Photo}\\
\cline{2-17}
&\multicolumn{4}{c|}{Short-term} & \multicolumn{4}{c|}{Long-term} & \multicolumn{4}{c|}{Short-term} & \multicolumn{4}{c|}{Long-term}\\
\cline{2-17}
   & 80ms & 160ms & 320ms & 400ms & 560ms & 640ms & 720ms & 1000ms & 80ms & 160ms & 320ms & 400ms & 560ms & 640ms & 720ms & 1000ms  \\
\hline
ERD\cite{fragkiadaki2015recurrent} & 1.13 & 1.20 & 1.59 & 1.78 & 1.86 & 2.03 & 2.09 & 2.59 & 0.70 & 0.78 & 0.97 & 1.09 & 1.20 & 1.23 & 1.27 & 1.37   \\
\hline
LSTM-3LR\cite{fragkiadaki2015recurrent} & 1.08 & 1.01 & 1.42 & 1.61 & 1.79 & 2.07 & 2.13 & 2.66 & 0.63 & 0.64 & 0.86 & 0.98 & 1.09 & 1.13 & 1.17 & 1.30 \\
\hline
Res-GRU\cite{martinez2017human} & 0.41 & 0.84 & 1.53 & 1.81 & 2.06 & 2.21 & 2.24 & 2.53 & 0.29 & 0.58 & 0.90 & 1.04 & 1.17 & 1.23 & 1.29 & 1.47 \\
\hline
Zero-velocity & \textbf{0.28} & \textbf{0.57} & \textbf{1.13} & \textbf{1.37} & \textbf{1.81} & 2.14 & 2.23 & 2.78 & \textbf{0.25} & \textbf{0.51} & \textbf{0.79} & \textbf{0.92} & \textbf{1.03} & \textbf{1.06} & \textbf{1.13} & \textbf{1.27} \\ 
\hline
Ours & 0.33 & 0.64 & 1.22 & 1.47 & 1.82 & \textbf{2.11} & \textbf{2.17} & \textbf{2.51} & 0.27 & 0.54 & 0.84 & 0.96 & 1.04 & 1.08 & 1.14 & 1.35 \\
\hline
 \end{tabular}
 }
     \label{Tab:SL_table}
     \vspace{-15pt}
\end{table*}

\subsection{Experimental Results}
We first evaluate the ability of the proposed framework for predicting human motion and made comparison with related recent state-of-the-art methods including ERD \cite{fragkiadaki2015recurrent}, LSTM-3LR \cite{jain2016structural} and Res-GRU \cite{martinez2017human}. We reproduced the results of these methods. Please note that our reproduced results often present better performance than that reported in their papers.
Following the evaluation of previous works, we converted the representation of joints from angle-axis to angle of rotation, and thereby measured the Euclidean distance between the predicted joints and its ground-truth by:
\begin{equation}
\begin{aligned}
&D(\{\mathbf{x}^t\}^T_{t=1}, \{\mathbf{\hat{x}}^t\}^T_{t=1}) = \\
             & \sum^T_{t=1} \sum^N_{i=1} \sqrt{d^2(\alpha_i^t, \hat{\alpha}_i^t) + d^2(\beta_i^t, \hat{\beta}_i^t) + d^2(\gamma_i^t, \hat{\gamma}_i^t)},
\end{aligned}
\end{equation}
where $d(a,b) = \text{min}\{|a-b|, 2\pi-|a-b| \}$.

Interestingly, the research in \cite{martinez2017human} found that repeating the last body pose also gave a relative small error in the measurement of the Euclidean distance between the ground-truth, which performed even better than ERD \cite{fragkiadaki2015recurrent} and LSTM-3LR \cite{fragkiadaki2015recurrent}. 
One possible reason is that the human motion within the dataset is slight for some activities. Therefore, simply repeating the last body pose can yield the reasonable objective results. Another possible reason may be attributed to the evaluation metric, which is an Euclidean distance and can only depict independent distance for each joint, and thus this ignores the relations between joints. Thus even with a smaller Euclidean distance, the motion prediction may not be plausible.

\begin{figure}[t]
\begin{center}
   \includegraphics[width=1\linewidth]{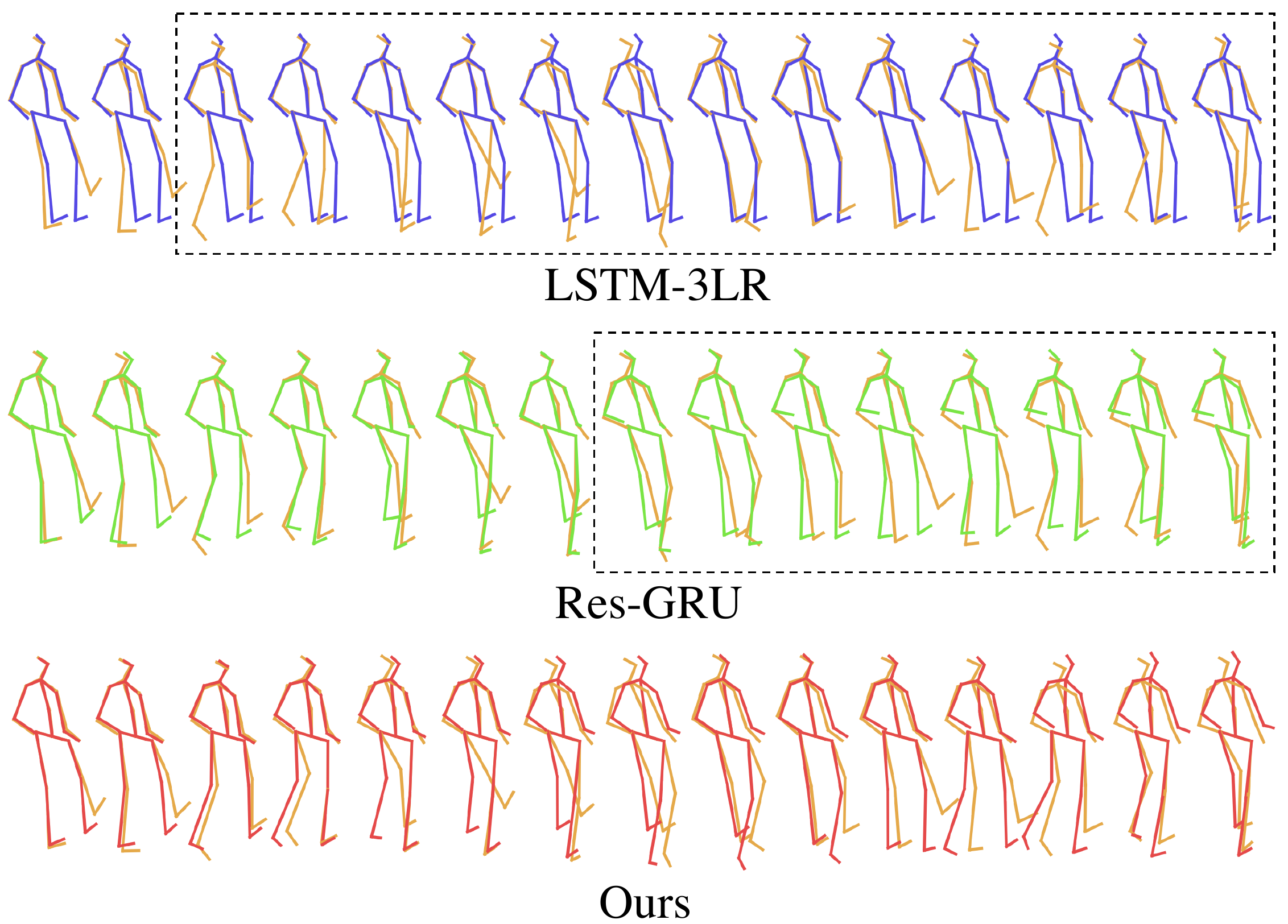}
\end{center}
\vspace{-11pt}
   \centering\caption{The comparison of mean pose convergence of the walking activity. The ground-truth poses are shown in yellow. The dash boxes highlight the converging motion sequence.}
\label{fig:compare}
\vspace{-11pt}
\end{figure}

We compare our results with the existing methods and the variants of our method. The first variant is using conventional mean square loss as in \cite{martinez2017human,fragkiadaki2015recurrent} and encoder-decoder framework with MHU.
We then replaced the MSE training loss with the gram matrix loss in the second variant. The fist and second variations are named as ``MHU-MSE'' and ``MHU-Gram'', respectively.

The overall human motion prediction result of all 15 activities of H3.6m dataset via mean angle error is shown in Table.\ref{Tab:all_table}. It can be observed that our methods as well as two different variants can outperform the competitors. More specifically, the result shows that MHU-MSE performs well in the short-term improvement of prediction especially from 160ms to 720ms. 
The reason can be attributed to that the MHU can efficiently filter motionless joints and propagate information between two layers with the modified non-linearity.
For the very short-term prediction, the spatial information of the body pose predominates the measurement. 
For longer term prediction, the motion information is more important.
Therefore, only considering the spatial pose information cannot well model the motion dynamic. As such MSE-Gram, which targets at enhancing motion dynamic, performs better, which achieves 1.82 on mean angle error at 1000ms. 
Finally, by assembling the motion context modeling, our model can achieve the best performance on both short-term and long-term predictions. 

In details, we show a part of the results in Table.~\ref{Tab:SL_table} which contains both short-term and long-term comparisons with the compared methods. 
In most of the cases, our results are competitive in short-term prediction, and clearly outperform the baseline methods in long-term prediction. 
For the ``walking'' activity, the objective evaluation of our result is close to that of Res-GRU. 
However, by visualizing the motion in Fig.~\ref{fig:compare}, it seems that the results generated by LSTM-3LR method and Res-GRU method converge to the mean body pose. 
On the contrary, our method can resemble the ``walking'' behaviors of the body pose, thus presenting the predicted motion with highly dynamic. 

\begin{figure}[t]
\begin{center}
   \includegraphics[width=1\linewidth]{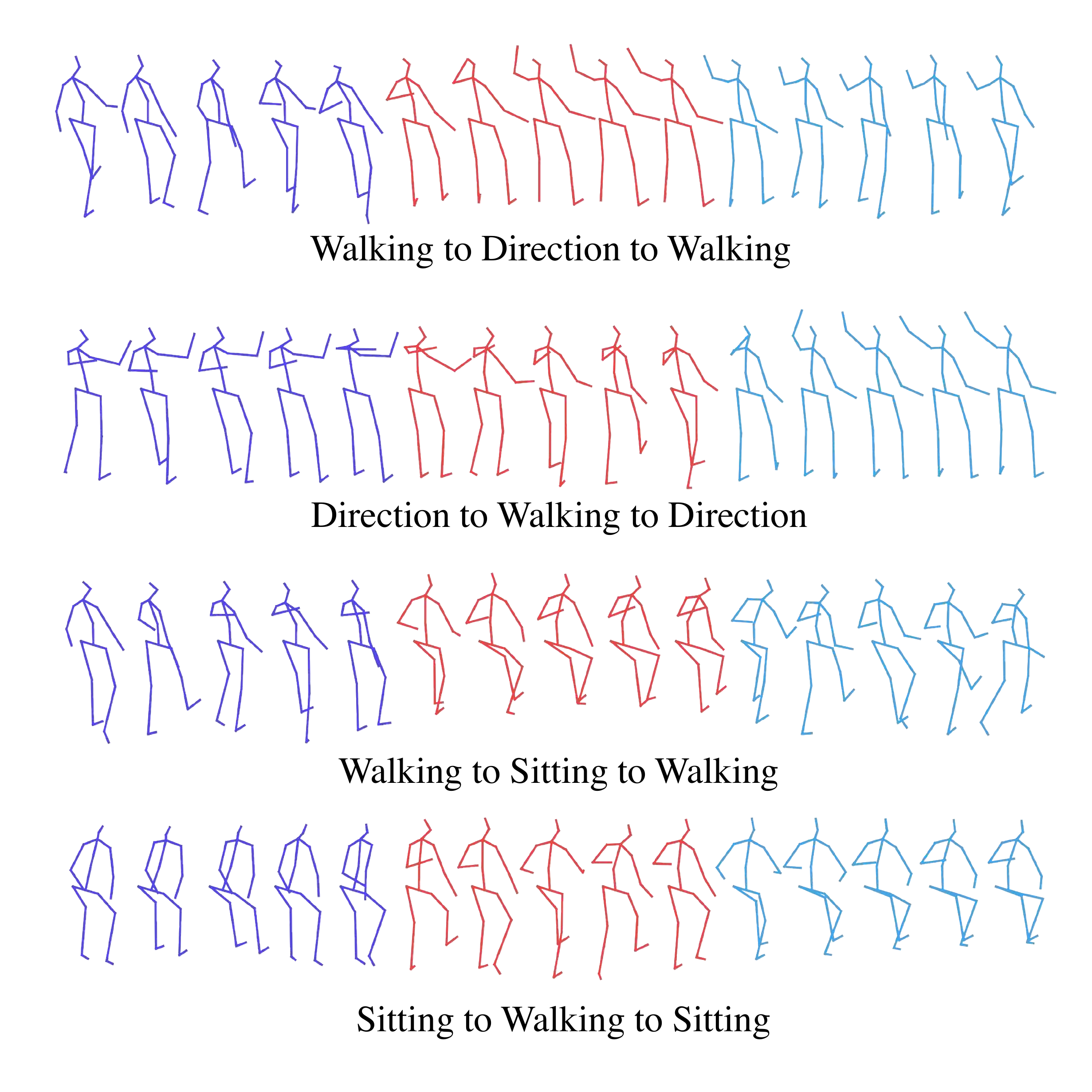}
\end{center}
\vspace{-21pt}
   \centering\caption{The result of human motion transfer between two different activities. The observed pose sequences are shown in purple, and the predicted motion are shown in red and blue. Better view in color.}
\label{fig:transfer}
\vspace{-11pt}
\end{figure}

\subsection{Human Motion Transfer}

As mentioned before, the long-term human motion is not deterministic and may alter according to subjective factors.
For example, while one is sitting, at any time he/she may suddenly stand up and walk around. Here, we try to simulate this situation by modifying the hidden state of each decoder input with the embedded activity label different from the observed activity. 

Since on the H3.6m dataset, there is no ground-truth human motion sequence that contains two different activities, we thus trained our model with the activity labels and feed a specific label at the test time. More specifically, after processing temporal attention procedure, we concatenated the hidden state with embedded action label, which was fed into a non-linear layer in order to construct the context representation for the decoding.

We show four examples of human motion transfer in Fig.~\ref{fig:transfer} including "walking" to "direction", "sitting" to "walking" and the inverses. By modifying the hidden states that encode the motion context of the subject, our method is able to transfer\renewcommand{\thefootnote}{\arabic{footnote}}\footnote{We use 'transfer' here because there is no ground-truth sequence to be 'predicted'.} human motion with smooth activity transitions. 
In details, as shown in the third row of Fig.~\ref{fig:transfer}, the proposed model manages to transfer human motion from "walking" to "sitting" and the inverse motion.
This result shows that the motion context encoded in the hidden states can be well extracted by the designed MHU to produce reliable human motion transfer.

\begin{figure}[t]
\begin{center}
   \includegraphics[width=0.85\linewidth]{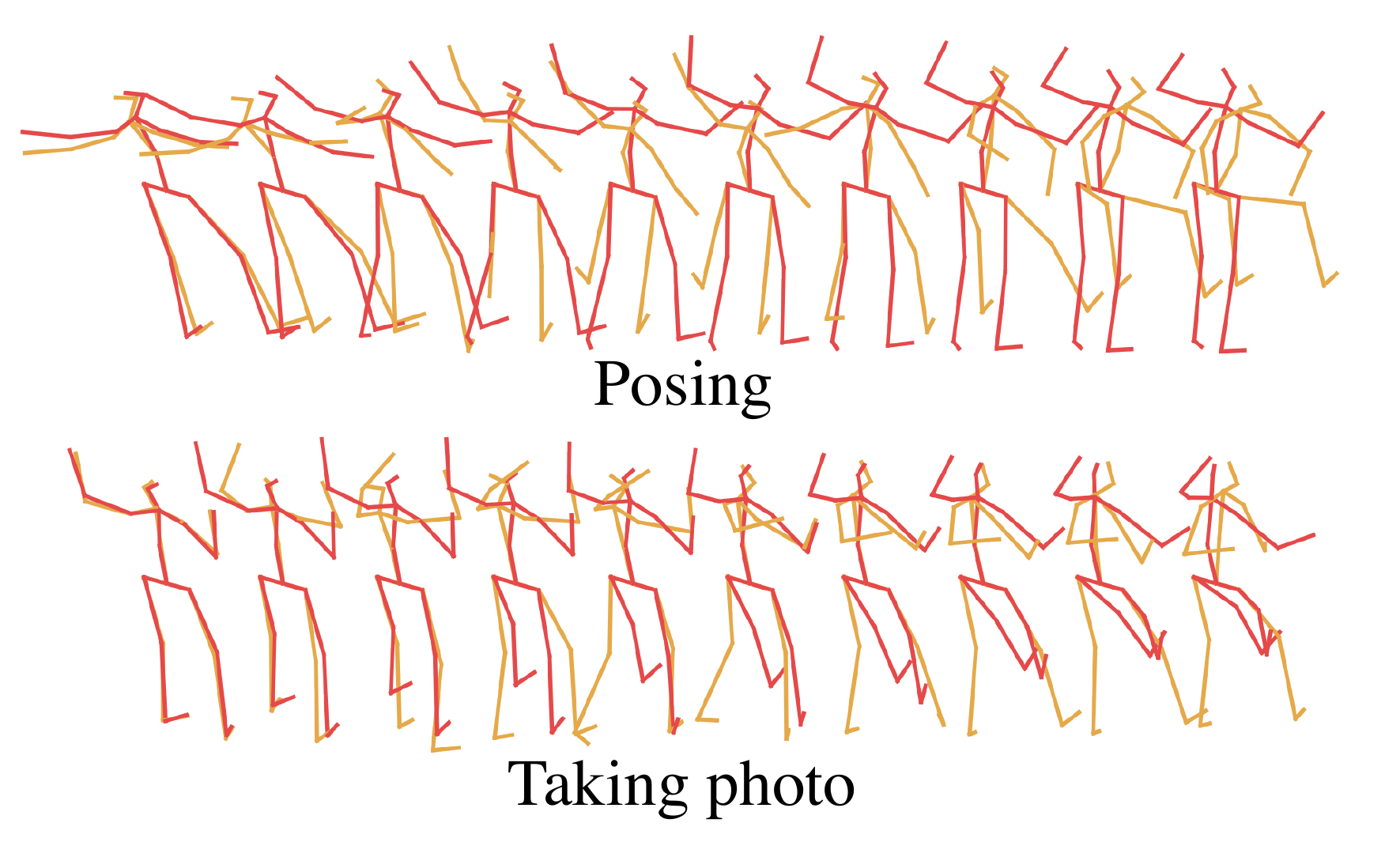}
\end{center}
\vspace{-15pt}
   \centering\caption{Failure cases of our human motion prediction method. The ground-truths and our results are shown in yellow and red, respectively.}
\vspace{-11pt}
\label{fig:lim}{}
\end{figure}

\subsection{Limitations}

Fig.~\ref{fig:lim} illustrates some failure cases of our method, which also happen for the existing methods. The main reason is that these activities are of high uncertainty with different subjects. Therefore, the observed information cannot provide enough evidence for modeling and predicting.

\section{Conclusion}


In this paper, we have proposed a new model to predict long-term human motions by exploring motion context and enhancing motion dynamic. 
The proposed motion context summarized the historical skeletons for providing fully observed evidence in long-term prediction. 
To enhance motion dynamic, the gram matrix training loss is further incorporated to capture the temporal transitions. The extensive results demonstrate that our proposed model outperforms existing methods especially for long-term motion prediction. Moreover, compared with other models, our model can perform human motion transfer which makes motion prediction based on the action command and alters the generated motion types accordingly.

\vspace{5pt}
\noindent\textbf{Acknowledgment} This work was supported partially by the National Key Research and Development Program of China (2018YFB1004903), NSFC(61522115, 61661130157), Guangdong Province Science and Technology Innovation Leading Talents (2016TX03X157), and the Royal Society Newton Advanced Fellowship (NA150459).

{\small
\bibliographystyle{named}
\bibliography{egbib}

\begin{thebibliography}{}

\bibitem[\protect\citeauthoryear{Aggarwal and Cai}{1997}]{aggarwal1997human}
Jake~K Aggarwal and Quin Cai.
\newblock Human motion analysis: A review.
\newblock In {\em Nonrigid and Articulated Motion Workshop, 1997. Proceedings.,
  IEEE}, pages 90--102. IEEE, 1997.

\bibitem[\protect\citeauthoryear{Bahdanau \bgroup \em et al.\egroup
  }{2014}]{bahdanau2014neural}
Dzmitry Bahdanau, Kyunghyun Cho, and Yoshua Bengio.
\newblock Neural machine translation by jointly learning to align and
  translate.
\newblock {\em arXiv preprint arXiv:1409.0473}, 2014.

\bibitem[\protect\citeauthoryear{Brand and Hertzmann}{2000}]{brand2000style}
Matthew Brand and Aaron Hertzmann.
\newblock Style machines.
\newblock In {\em Proceedings of the 27th annual conference on Computer
  graphics and interactive techniques}, pages 183--192. ACM
  Press/Addison-Wesley Publishing Co., 2000.

\bibitem[\protect\citeauthoryear{Cho \bgroup \em et al.\egroup
  }{2014}]{cho2014learning}
Kyunghyun Cho, Bart Van~Merri{\"e}nboer, Caglar Gulcehre, Dzmitry Bahdanau,
  Fethi Bougares, Holger Schwenk, and Yoshua Bengio.
\newblock Learning phrase representations using rnn encoder-decoder for
  statistical machine translation.
\newblock {\em arXiv preprint arXiv:1406.1078}, 2014.

\bibitem[\protect\citeauthoryear{Du \bgroup \em et al.\egroup
  }{2015}]{du2015hierarchical}
Yong Du, Wei Wang, and Liang Wang.
\newblock Hierarchical recurrent neural network for skeleton based action
  recognition.
\newblock In {\em Proceedings of the IEEE conference on computer vision and
  pattern recognition}, pages 1110--1118, 2015.

\bibitem[\protect\citeauthoryear{Fragkiadaki \bgroup \em et al.\egroup
  }{2015}]{fragkiadaki2015recurrent}
Katerina Fragkiadaki, Sergey Levine, Panna Felsen, and Jitendra Malik.
\newblock Recurrent network models for human dynamics.
\newblock In {\em Proceedings of the IEEE International Conference on Computer
  Vision}, pages 4346--4354, 2015.

\bibitem[\protect\citeauthoryear{Ghosh \bgroup \em et al.\egroup
  }{2017}]{Ghosh2017learning}
Partha Ghosh, Jie Song, Emre Aksan, and Otmar Hilliges.
\newblock Learning human motion models for long-term predictions.
\newblock {\em arXiv preprint arXiv:1704.02827}, 2017.

\bibitem[\protect\citeauthoryear{Grassia}{1998}]{grassia1998practical}
F~Sebastian Grassia.
\newblock Practical parameterization of rotations using the exponential map.
\newblock {\em Journal of graphics tools}, 3(3):29--48, 1998.

\bibitem[\protect\citeauthoryear{Hochreiter and
  Schmidhuber}{1997}]{hochreiter1997long}
Sepp Hochreiter and J{\"u}rgen Schmidhuber.
\newblock Long short-term memory.
\newblock {\em Neural computation}, 9(8):1735--1780, 1997.

\bibitem[\protect\citeauthoryear{Hu \bgroup \em et al.\egroup
  }{2015}]{hu2015jointly}
Jian-Fang Hu, Wei-Shi Zheng, Jianhuang Lai, and Jianguo Zhang.
\newblock Jointly learning heterogeneous features for rgb-d activity
  recognition.
\newblock In {\em Proceedings of the IEEE conference on computer vision and
  pattern recognition}, pages 5344--5352, 2015.

\bibitem[\protect\citeauthoryear{Huang \bgroup \em et al.\egroup
  }{2017}]{huang2017deep}
Zhiwu Huang, Chengde Wan, Thomas Probst, and Luc Van~Gool.
\newblock Deep learning on lie groups for skeleton-based action recognition.
\newblock In {\em Proceedings of the 2017 IEEE Conference on Computer Vision
  and Pattern Recognition (CVPR)}, pages 6099--6108. IEEE computer Society,
  2017.

\bibitem[\protect\citeauthoryear{Ionescu \bgroup \em et al.\egroup
  }{2014}]{h36m_pami}
Catalin Ionescu, Dragos Papava, Vlad Olaru, and Cristian Sminchisescu.
\newblock Human3.6m: Large scale datasets and predictive methods for 3d human
  sensing in natural environments.
\newblock {\em IEEE Transactions on Pattern Analysis and Machine Intelligence},
  36(7):1325--1339, jul 2014.

\bibitem[\protect\citeauthoryear{Jain \bgroup \em et al.\egroup
  }{2016}]{jain2016structural}
Ashesh Jain, Amir~R Zamir, Silvio Savarese, and Ashutosh Saxena.
\newblock Structural-rnn: Deep learning on spatio-temporal graphs.
\newblock In {\em Proceedings of the IEEE Conference on Computer Vision and
  Pattern Recognition}, pages 5308--5317, 2016.

\bibitem[\protect\citeauthoryear{Li and Chan}{2014}]{li20143d}
Sijin Li and Antoni~B Chan.
\newblock 3d human pose estimation from monocular images with deep
  convolutional neural network.
\newblock In {\em Asian Conference on Computer Vision}, pages 332--347.
  Springer, 2014.

\bibitem[\protect\citeauthoryear{Liu \bgroup \em et al.\egroup
  }{2016}]{liu2016spatio}
Jun Liu, Amir Shahroudy, Dong Xu, and Gang Wang.
\newblock Spatio-temporal lstm with trust gates for 3d human action
  recognition.
\newblock In {\em European Conference on Computer Vision}, pages 816--833.
  Springer, 2016.

\bibitem[\protect\citeauthoryear{Martinez \bgroup \em et al.\egroup
  }{2017}]{martinez2017human}
Julieta Martinez, Michael~J Black, and Javier Romero.
\newblock On human motion prediction using recurrent neural networks.
\newblock {\em arXiv preprint arXiv:1705.02445}, 2017.

\bibitem[\protect\citeauthoryear{Sutskever \bgroup \em et al.\egroup
  }{2014}]{sutskever2014sequence}
Ilya Sutskever, Oriol Vinyals, and Quoc~V Le.
\newblock Sequence to sequence learning with neural networks.
\newblock In {\em Advances in neural information processing systems}, pages
  3104--3112, 2014.

\bibitem[\protect\citeauthoryear{Taylor \bgroup \em et al.\egroup
  }{2007}]{taylor2007modeling}
Graham~W Taylor, Geoffrey~E Hinton, and Sam~T Roweis.
\newblock Modeling human motion using binary latent variables.
\newblock In {\em Advances in neural information processing systems}, pages
  1345--1352, 2007.

\bibitem[\protect\citeauthoryear{Tekin \bgroup \em et al.\egroup
  }{2017}]{tekin2017learning}
Bugra Tekin, Pablo Marquez~Neila, Mathieu Salzmann, and Pascal Fua.
\newblock Learning to fuse 2d and 3d image cues for monocular body pose
  estimation.
\newblock In {\em International Conference on Computer Vision (ICCV)}, number
  EPFL-CONF-230311, 2017.

\bibitem[\protect\citeauthoryear{Vemulapalli \bgroup \em et al.\egroup
  }{2014}]{vemulapalli2014human}
Raviteja Vemulapalli, Felipe Arrate, and Rama Chellappa.
\newblock Human action recognition by representing 3d skeletons as points in a
  lie group.
\newblock In {\em Proceedings of the IEEE conference on computer vision and
  pattern recognition}, pages 588--595, 2014.

\bibitem[\protect\citeauthoryear{Wang \bgroup \em et al.\egroup
  }{2014}]{wang2014learning}
Jiang Wang, Zicheng Liu, and Ying Wu.
\newblock Learning actionlet ensemble for 3d human action recognition.
\newblock In {\em Human Action Recognition with Depth Cameras}, pages 11--40.
  Springer, 2014.

\bibitem[\protect\citeauthoryear{Yasin \bgroup \em et al.\egroup
  }{2016}]{yasin2016dual}
Hashim Yasin, Umar Iqbal, Bjorn Kruger, Andreas Weber, and Juergen Gall.
\newblock A dual-source approach for 3d pose estimation from a single image.
\newblock In {\em Proceedings of the IEEE Conference on Computer Vision and
  Pattern Recognition}, pages 4948--4956, 2016.

\bibitem[\protect\citeauthoryear{Zilly \bgroup \em et al.\egroup
  }{2016}]{zilly2016recurrent}
Julian~Georg Zilly, Rupesh~Kumar Srivastava, Jan Koutn{\'\i}k, and J{\"u}rgen
  Schmidhuber.
\newblock Recurrent highway networks.
\newblock {\em arXiv preprint arXiv:1607.03474}, 2016.

\bibitem[\protect\citeauthoryear{Zimo \bgroup \em et al.\egroup
  }{2017}]{zimo2017auto}
Li~Zimo, Yi~Xiao, He~Chong, and Li~Hao.
\newblock Auto-conditioned lstm network for extended complex human motion
  synthesis.
\newblock {\em arXiv preprint arXiv:1707.05363}, 2017.

\end{thebibliography}
}

\end{document}